\documentclass{article} 
\usepackage{iclr2025_conference,times}

\usepackage{amsmath,amsfonts,bm}

\def\eqref#1{equation~\ref{#1}}

\def\1{\bm{1}}

\DeclareMathAlphabet{\mathsfit}{\encodingdefault}{\sfdefault}{m}{sl}
\SetMathAlphabet{\mathsfit}{bold}{\encodingdefault}{\sfdefault}{bx}{n}

\usepackage{hyperref}
\usepackage{url}
\usepackage{graphicx}

\title{Empowering LLMs in Decision Games through Algorithmic Data Synthesis}

\author{%
   Haolin Wang\textsuperscript{$1,2 \dag$},\quad 
   Xueyan Li\textsuperscript{$1 \dag$},\quad 
   Yazhe Niu\textsuperscript{$1,3$},\quad 
   Shuai Hu\textsuperscript{$1,4$},\quad 
   Hongsheng Li\textsuperscript{$3$}\\ 
   $^1$ Shanghai Artificial Intelligence Laboratory, $^2$ Beihang University\\
   $^3$ The Chinese University of Hong Kong, $^4$ Novosibirsk State University
}

\iclrfinalcopy 
\begin{document}

\maketitle

\begin{abstract}
Large Language Models (LLMs) have exhibited impressive capabilities across numerous domains, yet they often struggle with complex reasoning and decision-making tasks. 
Decision-making games, which inherently require multifaceted reasoning logic, serve as ideal sandboxes for evaluating and enhancing the reasoning abilities of LLMs. 
In this work, we first explore whether LLMs can master complex decision-making games through targeted post-training. 
To this end, we design data synthesis strategies and curate extensive offline datasets from two classic games, Doudizhu and Go. 
We further develop a suite of techniques to effectively incorporate this data into LLM training, resulting in two novel agents: Mastermind-Dou and Mastermind-Go. Our experimental results demonstrate that these Mastermind LLMs achieve competitive performance in their respective games. Additionally, we explore whether integrating decision-making data can enhance the general reasoning abilities of LLMs. Our findings suggest that such post-training improves certain aspects of reasoning, providing valuable insights for optimizing LLM data collection and synthesis strategies.
\end{abstract}

\section{Introduction}
\renewcommand{\thefootnote}{}
\footnotetext{$\dag$ Equal contribution. The dataset is available at:}
\footnotetext{ \texttt{\href{https://huggingface.co/datasets/OpenDILabCommunity/MasterMind}{https://huggingface.co/datasets/OpenDILabCommunity/MasterMind}}}
Language serves as an important role in human communication and the reflection of intricate thought processes. In recent years, the advancement of large language models (LLMs) has made it possible for artificial intelligence to understand and master human language \citep{brown2020language, chowdhery2023palm, touvron2023llama}, achieving human-level performances in various linguistic tasks \citep{wang2019superglue, adiwardana2020towards}. Although modern LLMs can be applied to a variety of tasks in a zero-shot manner, their logical reasoning abilities remain less than satisfactory, due to the absence of a comprehensive understanding of tasks at a deep and holistic level \citep{dziri2024faith}.

Research related to the reasoning capabilities of language models often focuses on code or mathematical problems. Researchers typically aim to evaluate and improve the reasoning capabilities of models on these tasks and have devised many approaches, including Self-Consistency (SC) \citep{wang2022self} and Chain-of-Thought (CoT) \citep{wei2022chain, kim2023cot}. However, while code and math serve as structured and symbolic data, it may not enclose the full reasoning process and lack textual format diversity expressed in natural language \citep{ma2023training}, which tends to be ambiguous. Consequently, adhering to the paradigm of evaluating and boosting LLMs with code and math problems may be not enough to generalize across varied tasks. A natural question arises: \textit{Is there another alternative tasks that can complement code and math, thereby better evaluating and elevating LLMs from the perspective of data quality and diversity?} In this work, we explore the potential of decision-making games as a complementary approach. We find that game data, when represented with appropriate textual structures, can provide valuable reasoning examples, broadening the scope for evaluating and enhancing language models.

\begin{figure}[!htbp]
\centering
\includegraphics[width=0.88\linewidth]{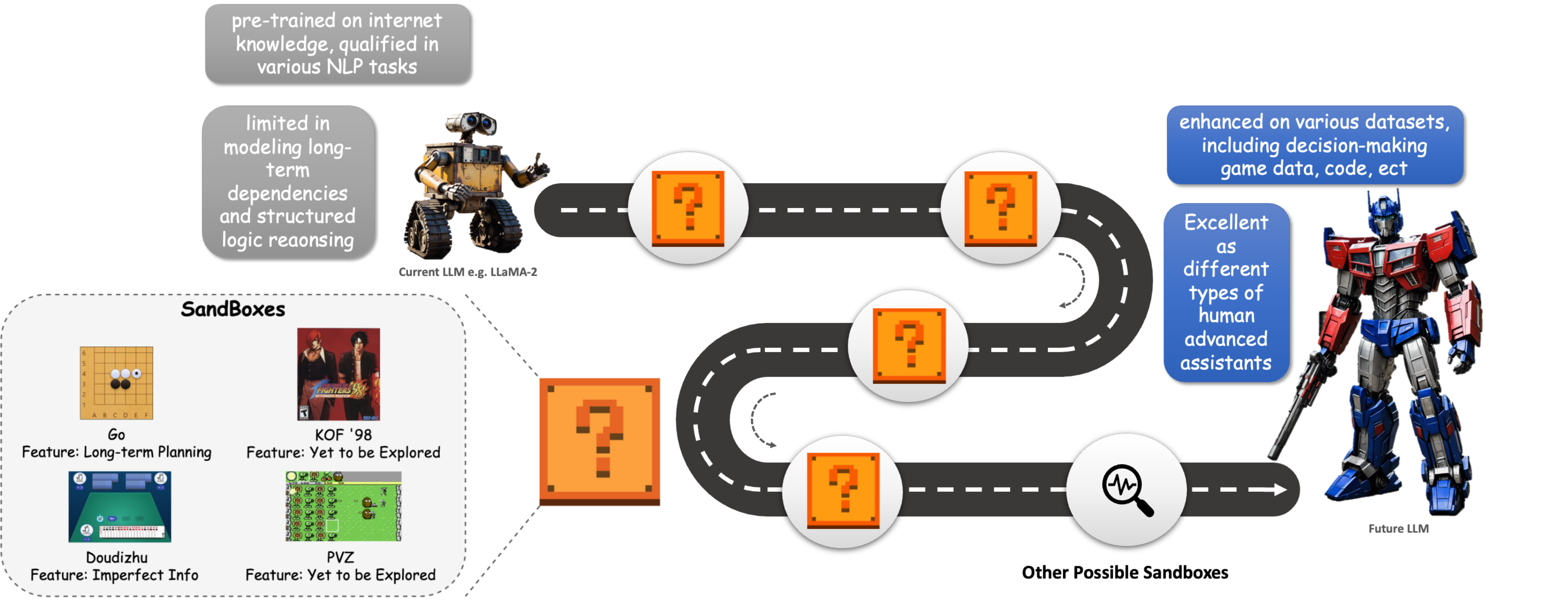}
\caption{By leveraging synthesized data from diverse decision-making games (such as Doudizhu and Go), current LLMs can be meticulously refined and enhanced, paving the way for their evolution into highly capable and intelligent agents in the future.}
\label{fig:mastermind_comic}
\end{figure}

Decision-making games, akin to code and math, usually posses a strict logical structure owing to its native mechanisms such as the state transition and the legal action range \citep{zha2019rlcard}. And the reasoning process in games can usually be split into several stages, simulating humans' step-by-step reasoning process. Moreover, games often contain complex strategies that may involve hierarchical planning \citep{levy2017hierarchical}, gambling operations \citep{silver2017mastering}, cooperative and competitive multi-agent behaviours \citep{lowe2017multi}, which can closely mirror reasoning strategies employed by humans. Additionally, responses from decision-making games under similar actions tend to exhibit consistency or follow a predictable distribution \citep{silver2018general}, fostering the stability.

Furthermore, game data exhibits a unique set of attributes. While code and math data often provides explicit information, game data include uncertainty and imperfect information, akin to real-world contexts. This imperfection can be further dynamically controlled when preparing training data for LLMs, as elucidated in the next section . Additionally, there are many available open-source agents on games \citep{niu2024lightzero}, enabling automated data collection across diverse environments. And a variety of game simulators \citep{brockman2016openai} allows for the customized generation of desired data. These properties not only diminish the requirements for manual annotation, but also enable the accumulation of larger training datasets for LLMs, which is easy to use for the development of more general agents. Moreover, many popular games like StarCraft \citep{starcraft_ii} and DOTA2 \citep{dota_2} necessitate multi-modal information, stochasticity modelling, and intricate game numerical design, which can therefore provide large models with multifaceted capability requirements.

In this paper, since we focus only on processing textual information, we conduct experiments on Doudizhu and Go, which challenge models with incomplete information and complex multi-step search capabilities, respectively. Our results show that current language models perform poorly on these tasks, even with few-shot learning and similarity based strategies. To address this, we aim to train models using tailored datasets to master these tasks effectively.

To implement, we initially collect datasets encompassing varying levels of strategies and game opponents. We meticulously implement some methods to convert the original game data to proper textual representations, such as the dynamic candidate actions and opponent responses in Doudizhu and the step-by-step natural language analysis in Go. Moreover, we design two techniques to prevent the overfitting of LLMs to complicated game rules, thereby ensuring a focus on the core reasoning process. To address the issue posed by the imperfect information in Doudizhu, we co-train an opponent strategy prediction model to anticipate the most probable actions of opponents in the next turn, providing more insights for LLMs' following selections. For the accuracy of the score difference calculation in Go, we integrate some tool functions (like code interpreter \citep{schick2024toolformer}). This integration serves to alleviate the cognitive burden on LLMs, enabling them to focus less on numerical details and more on developing a comprehensive game understanding.

By adopting the above data collection and training techniques, we fine-tune some open-source LLMs and enhance their game skills and reasoning abilities. We call derived models as Mastermind LLMs (agents). The experimental findings reveal that, our Mastermind LLMs not only reach competitive levels in corresponding games, but also display performance enhancements in unseen and general reasoning tasks. The main contributions can be summarized as follows:
\begin{itemize}
    \item We develop a series of techniques to fine-tune LLMs on two decision-making games, \textit{Doudizhu} and \textit{Go}, showcasing competitive performance in their respective tasks.
    \item We introduce decision-making games as a new data source for LLMs, highlighting their potential over code or math problems for enhancing step-by-step reasoning capabilities.
    \item We conduct thorough experiments to analyze the design of our data collection and training pipeline. All the code, data, and trained agents will be released soon.
\end{itemize}

\section{Related Work}
\subsection{LLM Data Preparation and Training}
The GPT model lineage has built the foundations of modern LLMs and achieved significant results on various tasks \citep{brown2020language, achiam2023gpt}.
These foundation language models are based on Transformer \citep{Vaswani+2017} architecture and adhere to the conventional "next token prediction" training objective employed on varied and massive text data.
These textual format data from various domains initially form the bedrock of knowledge for LLMs.
Recently, more data mixture strategies (e.g code) are designed to enhance LLMs both in the pre-trained stage \citep{touvron2023llama, team2024gemma} and the instruct tuning stage \citep{luo2023wizardcoder, li2024camel}.
Within a pre-defined symbolic and syntactic framework, code data can encapsulate various complicated process logics and data structures, enabling machines to execute them in a non-duality fashion. 
\citet{ma2023training} delved into the impact of code data on different training stages through empirical analysis.
\citet{yang2024if} provided a comprehensive account of several valuable attributes that emerge when code is integrated to training datasets of LLMs.
Also, a lot of open-sourced LLMs \citep{bai2023qwen} implied the pivotal role of code data in their technical reports, highlighting its significance in the evolution of language models.
Inspired by these seminal contributions, our research focus on the realm of data curation for LLMs.
And for the first time, we introduce decision-making games as an innovative and untapped data resource.

\subsection{Decision-making Games and Agents}
Markov Decision Process (MDP, $\mathcal{M}$) and Reinforcement Learning (RL) \citep{sutton2018reinforcement} stands as a foundational paradigm for decision-making problems, $\mathcal{M}$=$\left( \mathcal{S}, \mathcal{A}, \mathcal{P}, \mathcal{R},  \gamma, \rho_0\right)$, where $\mathcal{S}$ and $\mathcal{A}$ represent the state space and the action space.
And the transition function $\mathcal{P}$ assigns to each $(s, a)$ state-action pair a probability measure over $\mathcal{S}$.
Bolstered by the remarkable progress of deep learning~\citep{he2016deep}, deep RL have drawn many excellent achievements across different games \citep{vinyals2019grandmaster, schrittwieser2020mastering}.
More recently, there has been a surge interest in research community that regards LLMs as decision agents \citep{huang2022language,li2024camel}, aiming to utilize their internet-scale knowledge for execution of specific tasks.
To imbue LLMs with domain-specific knowledge, \citet{feng2024chessgpt} integrates both natural language and game interaction data in Chess, and employs contrastive learning to bridge policy training and language modeling.
The combination between LLM and RL agents can make it more consistent with human behavior, which may unlock many exciting opportunities such as a personalized tutor for game beginners.

\section{MasterMind-Dou}
Doudizhu \citep{doudizhu_wikipedia}, a popular Chinese card game played by three players, is well-suited for evaluating and enhancing the reasoning ability of LLMs due to its turn-based gameplay and controllable imperfect information. Players bid to become the 'landlord' and aim to empty their hands first, making strategic moves such as bluffing and reading opponents. When opponents' hands are revealed, decision-making shifts from gambling-based strategies to optimal action selection. Additionally, with a standard 54-card deck, the game's observation space and action space are easily convertible into text data, making it suitable for LLM training.

\subsection{Training Pipeline}
\begin{figure}[!tb]
\centering
\includegraphics[width=0.8\linewidth]{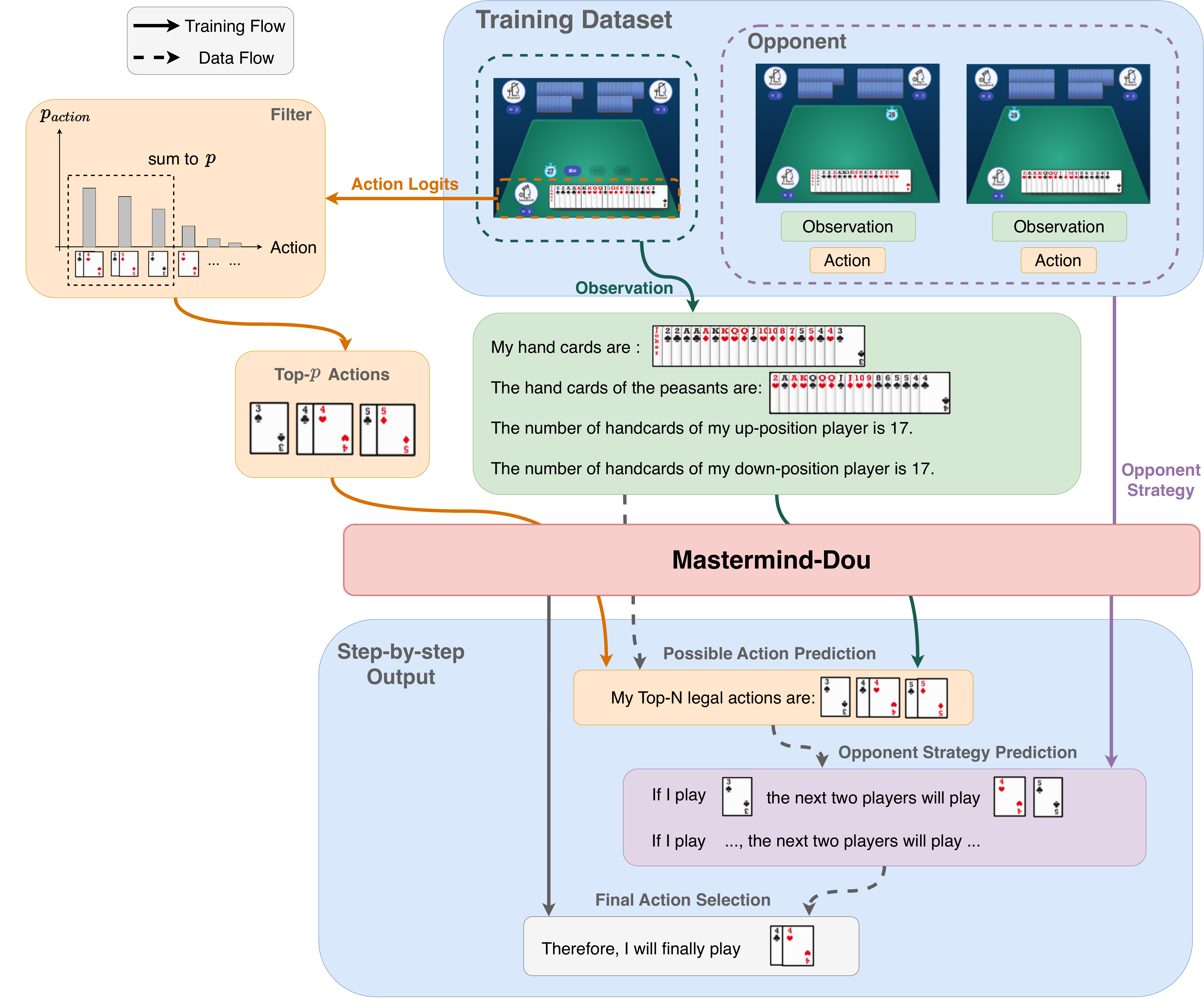}
\caption{MasterMind-Dou training pipeline workflow. The LLM first predicts probable legal moves, analyzes corresponding opponent responses, and then determines the optimal decision.}
\label{fig:doudizhu_pipeline}
\end{figure}

Doudizhu's outcome heavily influenced by specific hand cards and strategies, where optimal actions depend on both a player's observations and the unseen strategies of opponents. An example is provided in Appendix~\ref{appendix:dou_example}. To address this challenge, we propose a three-stage reasoning process, which includes possible action prediction, opponent strategy prediction, and final action selection (Fig.~\ref{fig:doudizhu_pipeline}). 
For detailed data definition and collection for each stage, please refer to Appendix~\ref{appendix:dou_dataset}

\textbf{\emph{Possible Action Prediction:}} Based on the textual representation of historical records and the player's current handcards, Mastermind-Dou is trained to predict the Top-$p$ possible actions associated with a heightened probability of winning.

\textbf{\emph{Opponent Strategy Prediction:}} To model unseen opponent strategies, we design an additional neural network head on the shelf of the LLM to predict the following action of opponents after the player performs a specific action. Specifically, we fine-tune the LLM to learn the common pattern among expert players, thereby enabling it to predict the most likely actions accurately. In addition, akin to in-context learning, we provide this fine-tuned model with samples of current opponents.


\textbf{\emph{Final Action Selection:}} Finally, by prognostically evaluating one's own possible actions opponents' reactions, we can concatenate these analysis to to construct the comprehensive query. Then the LLM will understand this information and generate the answer, namely, the final selected action.

\section{MasterMind-Go}
Complementing our exploration of Doudizhu, we extend our focus to Go as another domain to further enhance the reasoning ability of LLMs. In Go, black and white pieces are placed alternately on the board, with the ultimate goal of securing the largest possible territory. Compared to Doudizhu, Go presents greater complexity in both rules and strategy, demanding stronger logical reasoning skills. Additionally, since Go is played on a two-dimensional board while the model processes one-dimensional sequential input, LLMs must implicitly model spatial relationships. This sequential modeling challenge surpasses that of Doudizhu, adding another layer of difficulty.

To begin, we provide a brief overview of the reasoning process in Go. A typical game situation analysis involves several key steps: predicting the new state resulting from a given action, assessing the corresponding territory changes for both players, and evaluating the impact on the final win rate. A more detailed illustration is available in Appendix~\ref{appendix:go-rule}.  

Given that playing Go requires various types of logical reasoning skills, we design a curriculum of training tasks that follow a gradual, step-by-step progression. Our approach emphasizes embedding comprehensive thought processes into the LLM’s cognitive framework for gameplay. This principle aligns with procedure cloning \citep{wei2022chain}, allowing the model to internalize the underlying logic of decision-making rather than merely mapping states to actions. Additionally, this method resonates with the concept of Chain of Thought (CoT), encouraging the model to approach problem-solving in a structured, stepwise manner.

\begin{figure}[!tp]
\centering
\includegraphics[width=0.75\linewidth]{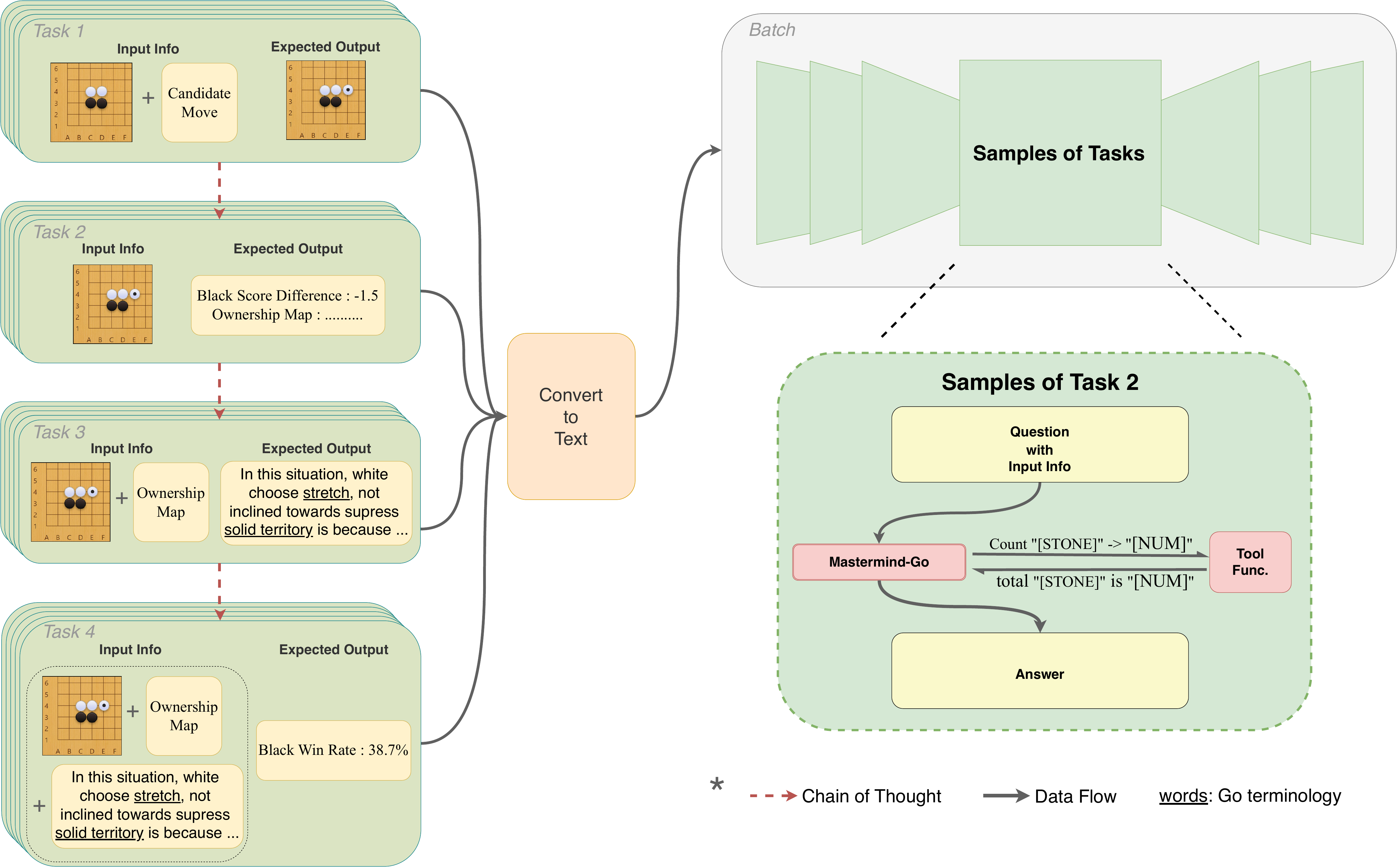}
\caption{The training pipeline overview of Mastermind-Go. It is designed to incorporate a hierarchical approach, featuring four distinct tasks that generate progressive and various Go board analyses. These analyses are then transformed into textual data for fine-tuning the LLM.}
\label{fig:go_pipeline}
\end{figure}

Specifically, we divide the reasoning process of Go into four stages and collect datasets for each corresponding task, as illustrated in Fig.~\ref{fig:go_pipeline}. Task 1 is from the rule level, where we have the model predict the next state based on the current board state and actions. Task 2 is from short-term and state understanding level, where we use the open-source Go agent, KataGo, to generate evaluations for current board. Task 3 is from long-term and natural language understanding level, where we utilize information from Go books to extract several state-explanation pairs, thereby enhancing the model's understanding of the game states through natural language. Task 4 is from the decision level, where we combine all information mentioned above to make the final decision in a step by step manner. We detail the specific methods of data collection in Appendix~\ref{appendix:go_dataset}

\subsection{Training and Inference Pipeline}
After data collections across different tasks, we mix up these samples with calibrated proportions to fine-tune the LLM. Within each training instance, a question-answer pair is presented. We denote the question of the $i$-th sample as $X_i$ and the answer as $Y_i$. Our training regimen employs the standard SFT loss function, which maximize the likelihood probability of yielding the complate answer upon receiving an input query. Formally, this loss function can be defined as: $L_{\theta} = \frac{1}{N}\sum_{i}^{N} \log_{\theta} (Y_i \| X_i).$


However, we find that the LLM performs poorly in counting the territories of both black and white when given an ownership map, making the subsequent prediction of winning probabilities harder. To address this issue and enhance performance, we design a "Count" function that allows the LLM to programmatically calculate territories while generating answers, leading to more reliable results.

\section{Experiments}
Next, we empirically explore the impact of fine-tuning LLMs on the generated decision-making game data. In terms of experimental setup, we selecte the LLaMA-2-7B model as the primary base model for training, with the main training conducted on a machine equipped with 8$\times$A100 GPUs. Detailed training hyper-parameters are shown in Appendix~\ref{appendix:hyper}. For more experimental results and the effect of training on general reasoning abilities of LLMs, please refer to Appendix~\ref{appendix:exp_addition}.

\subsection{Doudizhu Results}
To assess the performance of Doudizhu tasks, our evaluation concentrates on three primary metrics: alignment with expert data via action accuracy (\textbf{Act. Acc.}) and thought process matching (\textbf{CoT RL}), similarity (Rouge-L) in forecasting the opponent's card-play probability (\textbf{Pred. Acc}). The results are listed in the following table.
To facilitate a better comparison, we evaluated five different methods: \textbf{LLaMA-2-7B (0-shot):} Directly inputting the question into the model, allowing the model to generate subsequent thoughts and responses, predicting the final move strategy. \textbf{LLaMA-2-7B (few-shot):} Providing several examples to the model as context, prompting the model to respond in a similar manner and give the final move strategy. \textbf{LLaMA-2-7B (few-shot + similarity):} Selecting the most relevant action from all possible actions by calculating the similarity of embeddings \citep{huang2022language}, and using it as the final move strategy. \textbf{Mastermind-Dou w/o prob:} Inputting the question into the model, which directly provides the final answer without intermediate reasoning. \textbf{Mastermind-Dou with prob:} Inputting the question into the model, allowing the model to perform step-by-step intermediate reasoning and then provide a comprehensive final answer.

\begin{table}[h]
    \centering
    \begin{tabular}{l|c|c|c}
        \hline
        & \textbf{Act. Acc. $\uparrow$} & \textbf{CoT. RL $\uparrow$} & \textbf{Pred. Acc. $\uparrow$} \\
        \hline
        LLaMA-2-7B (0-shot) & 0\% & 0.52 & 0\% \\
        LLaMA-2-7B (few-shot) & 21.18\% & N/A & N/A \\
        LLaMA-2-7B (few-shot + similarity) & 42.65\% & N/A & N/A \\
        Mastermind-Dou w/o prob & 78\% & N/A & N/A \\
        Mastermind-Dou with prob & \textbf{90\%} & \textbf{0.98} & \textbf{39\%} \\
        \hline
    \end{tabular}
    \caption{Performance comparison of different models on Doudizhu tasks.}
    \label{tab:model_performance}
\end{table}

Firstly, the performance on the Doudizhu task, as shown in the Table~\ref{tab:model_performance}, demonstrates that the model before training is not able to grasp the rule of such a card game. After training, the model has clearly acquired card-playing abilities and exhibits strong generalization capabilities. On previously unseen situations, the model achieves an 90\% match with expert data and demonstrates a 98\% similarity in thought processes. This indicates that the language model has not merely memorized specific card-playing strategies for certain scenarios but has learned an intrinsic logic of the Doudizhu game, showcasing excellent generalizability. 

Additionally, comparing the results with and without Chain-of-Thought (CoT) training reveals that incorporating a thought process can enhance the model's ability to generalize in its final decision-making. It is worth noting, however, that although the overall similarity in thought process is high, achieving perfect prediction of the entire thought chain is challenging, with an accuracy rate of only around 39\%. This is likely due to Doudizhu's characteristic of imperfect information, where neither side can fully know the opponent’s hand, making it difficult to perfectly replicate the correct sequence of predictions about the opponent’s moves.

\begin{table}[h]
    \centering
    \begin{tabular}{l|c|c}
        \hline
        & \textbf{v.s. RLCard $\uparrow$} & \textbf{v.s. DouZero $\uparrow$} \\
        \hline
        LLaMA-2-7B & 0\% & 0\% \\
        LLaMA-2-7B (few-shot) & 12\% & 3\% \\
        Mastermind-Dou w/o prob & 78\% & 33\% \\
        Mastermind-Dou with prob & \textbf{90\%} & \textbf{41\%} \\
        DouZero (Expert) & 90\% & 43\% \\
        \hline
    \end{tabular}
    \caption{Win rates of different LLMs against RLCard and DouZero on the Doudizhu task.}
    \label{tab:vs_rlcard_douzero}
\end{table}

We further evaluate the trained agent by conducting match-ups with various open-source solutions, repeating each round 100 times. To ensure fair testing and eliminate the influence of teammate capabilities, our proposed agent was assigned the "landlord" role, while the two "farmer" roles were consistently filled by open-source agents. The results, as shown in Table~\ref{tab:vs_rlcard_douzero}, indicate that our proposed agent demonstrates strong performances. Against RLCard \citep{zha2019rlcard} as the opponent, our agent achieves a 90\% win rate, effectively inheriting the expertise-level performance. Similarly, when pitted against the expert model DouZero \citep{zha2021douzero} it achieved a 41\% win rate, which is very close to DouZero's own expert-level win rate limit of 43\%.

Interestingly, analysis reveals that our agent is not merely imitating DouZero’s actions; its strategy is not always aligned with DouZero’s. In some situations, despite DouZero ultimately losing, our agent—using a different strategy—achieved victory. This suggests that the language model may indeed be learning certain high-level strategies rather than simply performing behavior cloning. An example can be viewed in Appendix~\ref{appendix:case_study}.

\subsection{Go Results}
Due to the highly complex nature of Go, with its vast array of variations and its two-dimensional spatial input, the game is not ideally suited for language model learning. After initial attempts, we find that teaching a language model through supervised learning in this context incurs significant costs. Hence, we only evaluate Mastermind’s performance in Go on a few proxy tasks. 

Firstly, from the perspective of rule comprehension, we test the prediction accuracy for next board states (\textbf{s' Acc.}). Secondly, to assess understanding of the game state, we measured the mean absolute error in score differences (\textbf{Score MAE}) and win rates (\textbf{Winrate MAE}). Lastly, on the natural language interpretation level, we examine the Rouge-L similarity between generated explanations and actual explanations (\textbf{expl. RL}), as well as the perplexity of true explanations (\textbf{expl. ppl.}).

\begin{table}[h]
    \centering
    \begin{tabular}{l|c|c|c|c|c}
        \hline
        & \textbf{s' Acc. $\uparrow$} & \textbf{Score MAE $\downarrow$} & \textbf{Winrate MAE $\downarrow$} & \textbf{expl. RL $\uparrow$} & \textbf{expl. ppl. $\downarrow$} \\
        \hline
        LLaMA-2-7B & 0\% & N/A & N/A & 0.28 & 11.45 \\
        Single-task & \textbf{99.44\%} & 1.80 & 5.14 & \textbf{0.44} & 5.23 \\
        Multi-task & 96.08\% & \textbf{1.74} & \textbf{4.49} & 0.43 & \textbf{3.64} \\
        \hline
    \end{tabular}
    \caption{Performance comparison of different models on Go tasks.}
    \label{tab:performance_metrics}
\end{table}

From Table~\ref{tab:performance_metrics}, it is evident that the MasterMind agent has developed a strong understanding of Go’s rules, achieving nearly flawless accuracy in predicting the next game state. This is particularly notable given the inherent complexity of Go, where a single move can significantly alter the status of a large board area.  
Moreover, we find that the model excels in analyzing more challenging game situations and accurately estimating the territorial scope for both players. Additionally, MasterMind's win rate predictions exhibit high accuracy, with variance kept within 5\%. In Appendix~\ref{appendix:case_study}, we provide examples demonstrating MasterMind's accuracy in territory estimation and win rate prediction.

\subsection{General Reasoning Tasks}
Additionally, we investigate the impact of training on decision-making games on the model's general reasoning ability. We validate Mastermind LLMs on BIG-Bench Hard (BBH) \citep{suzgun2022challenging}, a challenging reasoning dataset for LLM.

\begin{table}[h]
    \centering
    \begin{tabular}{l|c|c|c|c|c|c}
        \hline
        \textbf{Models} & \textbf{TempSeq} & \textbf{PengTab} & \textbf{Snarks} & \textbf{RuinNames} & \textbf{Hyper.} & \textbf{Nav.} \\
        \hline
        LLaMA-2-7B & 12.00\% & 31.51\% & 47.75\% & 32.80\% & 51.60\% & 53.60\% \\
        Mastermind-Dou & 20.00\% & \textbf{35.62\%} & 49.44\% & 35.60\% & \textbf{62.80\%} & 56.80\% \\
        Mastermind-Go & \textbf{20.40\%} & 29.45\% & \textbf{51.69\%} & \textbf{39.20\%} & 51.60\% & \textbf{60.00\%} \\
        \hline
    \end{tabular}
    \caption{Accuracy improvements across different tasks in BBH.}
    \label{tab:accuracy_comparison}
\end{table}

Firstly, we observe a significant performance improvement in the model on specific subsets following MasterMind training, as shown in Table~\ref{tab:accuracy_comparison}. This improvement appears most prominently in reasoning tasks that require long-sequence modeling, likely due to the demands of both Go and Doudizhu, which test the model’s ability to make accurate predictions over extended outputs. 

\begin{table}[h]
    \centering
    \begin{tabular}{l|c|c|c|c}
        \hline
        \textbf{Models} & \textbf{Disambiguation QA} & \textbf{Date} & \textbf{Geometric} & \textbf{Three Objects} \\
        \hline
        LLaMA-2-7B & \textbf{46.8\%} & \textbf{60.4\%} & \textbf{19.2\%} & \textbf{55.2} \\
        Mastermind-Dou & 32.4\% & 22.8\% & 0\% & 41.6 \\
        Mastermind-Go & 44.8\% & 57.6\% & 7.6\% & 51.2 \\
        \hline
    \end{tabular}
    \caption{Accuracy decline across different reasoning tasks in BBH.}
    \label{tab:accuracy_reasoning}
\end{table}

However, it is important to note that the model exhibits a decline in performance on certain tasks, as highlighted in Table~\ref{tab:accuracy_reasoning}. This decline is likely attributable to catastrophic forgetting, a phenomenon where certain skills or knowledge, such as information about dates and spatial figures, are not adequately reinforced during the MasterMind training process.

To fully address this issue, we believe that introducing these data into the pretraining phase may be necessary to enhance overall performance. However, due to resource constraints, we are unable to implement this idea in the current study, leaving it as a potential direction for future work.

\subsection{Training with Different Models}

We also conduct additional ablations on other types of LLMs, selecting Google’s Gemma \citep{team2024gemma} and fine-tuning it on the Doudizhu dataset. As shown in Table~\ref{tab:gemma_performance}, similar improvements are observed in the Doudizhu-related metrics across different types and sizes of LLMs. Furthermore, the overall performance gains are more pronounced in larger models, indicating a scaling law effect. This experiment supports the idea that decision-making game data could serve as a valuable data source for open-source LLMs in general.

\begin{table}[h]
    \centering
    \begin{tabular}{l|c|c}
        \hline
        \textbf{Model} & \textbf{Act. Acc. $\uparrow$} & \textbf{CoT RLsum $\uparrow$} \\
        \hline
        Gemma-2B & 0\% & 0.40 \\
        Gemma-7B & 0\% & 0.44 \\
        Gemma-2B-Mastermind & 76.69\% & 0.97 \\
        \textbf{Gemma-7B-Mastermind} & \textbf{86.27\%} & \textbf{0.98} \\
        \hline
    \end{tabular}
    \caption{Performance comparison of Gemma models on the Doudizhu task.}
    \label{tab:gemma_performance}
\end{table}

\section{Conclusion}
In this paper, we delve into empowering LLMs for mastering complex reasoning games. Unlike many existing works that focus on code and math problems, we incorporate decision-making games to evaluate and boost the reasoning capacity of LLMs. Through a suite of our designed techniques in data collection and training, we have developed MasterMind agents, demonstrating commendable performance in both Doudizhu and Go. Empirical experiments also serve to substantiate the potential of this approach in improving general reasoning capabilities of LLMs.

\bibliography{iclr2025_conference}
\bibliographystyle{iclr2025_conference}
\clearpage

\appendix

\section{Dataset Collection}
\subsection{Doudizhu Dataset Collection}
\label{appendix:dou_dataset}

\emph{Card Textual Representation.}
To simply and clearly represent the data in Doudizhu, we have chosen to build upon and refine the approach introduced in DouZero \citep{zha2021douzero}. Specifically, each card is translated into an integer, and possible combinations of cards are represented as sorted lists of integers. The detailed correspondences are listed in  Table~\ref{appendix:tab_card_encode}.

\renewcommand{\arraystretch}{1.3} 
\begin{table}[h]
\centering
\begin{tabular}{cc||cc||cc||cc}
\hline
\textbf{Card Face} & \textbf{Integer} & \textbf{Card Face} & \textbf{Integer} & \textbf{Card Face} & \textbf{Integer} & \textbf{Card Face} & \textbf{Integer} \\ \hline
3                  & 3                              & 4                  & 4                              & 5                  & 5                              & 6                  & 6                              \\ 
7                  & 7                              & 8                  & 8                              & 9                  & 9                              & 10                 & 10                             \\ 
J                  & 11                             & Q                  & 12                             & K                  & 13                             & A                  & 14                             \\ 
2                  & 17                             & Black Joker        & 20                             & Red Joker          & 30                             &                    &                                \\ \hline
\end{tabular}
\caption{Card to Integer Mapping. A card with higher ranking corresponds to a larger integer value.}
\label{appendix:tab_card_encode}
\end{table}

\emph{Preliminary Filtering of Candidate Actions.}
We discretize and simplify the action space by providing a list of legal actions, a critical step in reducing the complexity of the space that LLM need to consider when making decisions. However, the extensive legal action space in Doudizhu, exceeding 27,000 combinations, poses a challenge in context length and reasoning complexity for LLM. Furthermore, the fluctuating length of legal actions poses a significant obstacle for LLMs to develop adaptable strategies. To tackle this, we utilize a pre-trained Q network from DouZero to obtain the logits for each action and employ Top-$p$ sampling, choosing actions with a combined probability of 25\%, thus reducing the action space size while preserving the optimal action sets for winning.

\emph{Data Generating Agent.}
We require diverse datasets of different player strategies across distinct positions.  This is essential as, in any given position, we need to predict not only the optimal action set for the current player but also opponents' potential moves. Furthermore, the network we intend to train need to generalize to any new opponent strategies, underscoring the need for a varied data source. Therefore, we utilize three types of agents: rule-based \citep{zha2019rlcard}, supervised learning-based, and DouZero agents \citep{zha2021douzero} enhanced by RL. Each agent contributes to generating strategy datasets by providing prompts that specify their proficiency levels, mapping states to Top-$N$ actions ($s\rightarrow a$).

\emph{Detailed Prompt Templates.}
We next detail the prompt templates utilized in each task for curating Doudizhu dataset as follows:

\textbf{Possible action prediction:}
\begin{equation}
\mbox{\textit{My handcards are ..., while the opponents' handcards are ...}. \textit{My possible actions are ...} }. \notag
\end{equation}

\textbf{Opponent strategy prediction:}
\begin{align*}
&\mbox{\textit{My handcards are ..., while the opponents' handcards are ... }. \textit{My action is ...}}  \notag
\end{align*}

\textbf{Final action selection:}
\begin{equation}
 \mbox{\textit{If I play a, the next two players will play} } \mbox{\textit{a}}_1, \mbox{\textit{a}}_2... \notag \\
 \mbox{\textit{Therefore, I will play a}}.
\end{equation}

\subsection{Go Dataset Collection}
\label{appendix:go_dataset}

\emph{Board Textual Representation:} Firstly, we explain the method of converting the original Go board into its textual form.  According to the rules of Go, the dimensions of board is fixed at 19x19, with each slot potentially occupied by a black stone, a white stone, or a remaining vacant. Since the board is 2-dimensional, it is challenging to convert it into 1-dimensional language sequence without losing any spatial information. In our settings, we use symbols "o", "\#" and "•" to represent white stones, black stones, and unoccupied positions. To facilitate precise reference to each position on the board, we assign labels to 19 rows and 19 columns, ranging from 1 to 19 and A to T, respectively, with the position at the bottom-left corner denoted as "A1". Additionally, to encode historical information, we assign numbers to the most recent moves of both sides, such as "o(1)" and "\#(3)" to denote the first move by a white stone and the third move by a black stone.  This approach completes the representation of the board, converting all necessary information on the board into textual format. 

\emph{Data Collection for Basic Rules:} To ensure the model grasps the core rules of Go (especially capturing stones), we begin with predicting the state-action transition of the game.  Typically, there are two types of pre-training tasks to achieve this goal: 1) $s,s'\rightarrow a$, i.e. predicting corresponding action given adjacent board states; 2) $s,a\rightarrow s'$, i.e. predicting next board state given current state and action. Ultimately, the collected data is presented as follows:

\begin{align*}
\mbox{\textit{The}} &\mbox{\textit{ following is a game record of 19x19 Go. ... The next move is on ... }} \\
&\mbox{\textit{Please predict the next Go board after this move ...}} \notag
\end{align*}

\emph{Data Collection from Agents:} After understanding the fundamental rules of Go, we choose KataGo \citep{wu2019accelerating} as the agent for generating evaluations about current game states. KataGo is a RL agent based on Monte Carlo Tree Search (MCTS), displaying capabilities that exceed human-level in Go. 

In building our dataset, we didn't mimic MCTS search paths due to the unknown logic behind Go moves in MCTS and the exponential growth in search space, which lowers efficiency with too many tokens. Our approach aims to help the model grasp the reasoning behind move placements, implemented as follows: Firstly, we record KataGo self-play games and save the state of the board at each move. Next, we analyze each board state to identify several potential move candidates. After obtaining these potential moves, we simulate the board state after playing these moves and passed the updated board state to KataGo for analysis. We include three aspects of analysis information on the simulation result: 1) the ownership of territories on the board; 2) the lead in points for current player; 3) the winning probability in the current board state. Finally, all the above information is converted into text form and stored in the training data. 

In crafting the model's response, we have adopted a similar design to the CoT approach. Initially, the model is guided to predict the basic ownership map, subsequently gauging the respective point difference based on the size of territories owned by black and white sides, respectively. Finally, it predicts the winning probability. By employing this progressive thinking method, the LLM is able to better utilize decision-making data for reasoning. The training samples finally adopt this template:

\begin{align*}
\mbox{\textit{The}}\mbox{\textit{ following is a game board of Go. ... Here is the board state: ...}} \\
\mbox{\textit{Please predict the ownership map ... leading score ... and win rate.}} \notag
\end{align*}

\emph{Data Collection from Books:} The next step is to enable the model to accurately assess the current state on from a long-term and natural language perspective. 
Thus, we leverage resources from existing Go books \citep{LeeSedol}, extracting the pairs of game states and corresponding explanations about judgement of current situation and long-term strategy.
The training sample thus adopts the following format:

\begin{align*}
\mbox{\textit{The}} &\mbox{\textit{ following is a game board of Go. ... Here is the board state: ...}} \\
&\mbox{\textit{Please generate the corresponding explanations. ...}} \notag
\end{align*}

\emph{Combined Training:} Finally, we combine the above data for training as a complete task to boost the LLM. In doing so, Mastermind-Go fully utilizes Go training data from various levels of thought, integrating all knowledge to deliver the optimal decision.

To enrich our dataset with more diversity, we choose two distinct levels of agents for data-generation. The first one is called KataGo-9d, an agent that consistently employs the optimal move; the second one is KataGo-suboptimal, which uses the Top-$p$ sampling to randomly stochastically actions according to the policy network in Katago. We choose $p=0.4$ in our implementation.
This dual-tiered approach facilitates comprehensive game trajectories.

\subsection{Overview of Datasets Information}
\label{appendix:dataset_info}
We have placed the basic information of the dataset collection in Table 5 below. For the Doudizhu task, there are three related datasets: the dataset without card-playing probability thought chains (Dou-no-prob), the dataset with card-playing probability thought chains (Dou-prob), and the dataset for predicting the opponent's card-playing probability (Dou-pred-prob). For the Go task, there are also three corresponding datasets: the dataset for predicting environment dynamics (Go-next-state), the KataGo position analysis dataset (Go-analysis), and the natural language analysis dataset of state-explanation pairs (Go-state-expl).

Regarding the evaluation metrics for the performance of Doudizhu datasets, there are generally three aspects. The first aspect is the alignment with expert data, including the accuracy of the final selected action (namely \textbf{Act. Acc.}), and the degree of match between the LLM's thought process and the expert's thought process. We referred to the method of \citet{lin2004rouge} to measure this similarity (namely \textbf{CoT RLsum}). The second aspect is the accuracy of predicting the opponent's card-playing probability, we choose the final prediction accuracy (namely \textbf{Pred. Acc.}) as metirc. The third aspect is the win rate (namely \textbf{Win-Rate}) against the RLcard strategy.

For metrics to evaluate the performance of Go datasets, we have set different configurations for various tasks, as detailed below. For task 1, which involves predicting the next board state, we use the accuracy of prediction (namely \textbf{s' Acc.})  as the evaluation result. If the prediction for all positions of stones on a sample is correct, the sample is marked as correct. For task 2, we used the mean absolute error (MAE) of the difference in points and win rates from the real values (namely \textbf{Score MAE} and \textbf{Win-Rate MAE}, respectively). If a sample does not generate a correct format for score difference or win rate, the errors are respectively set to 10 and 100\% for subsequent average calculation. For task 3, which involves predicting the explanation of the game state, we evaluate the similarity between generated content and label explanation using rouge metric (namely \textbf{expl. RLsum}). The higher this value, the more similar the generated results are to the target. We also present the perplexity of ground truth explanations (namely \textbf{expl. ppl.}) 

\begin{table}[h]
\centering
\begin{tabular}{lrrrl}
\hline
\textbf{Dataset}            & \textbf{\# Trajectories} & \textbf{\# Data Samples} & \textbf{\# Tokens} & \textbf{Metrics}                 \\ \hline
Dou-prob            & 1,500                                         & 1,712,802                                     & 0.78 T                                  & Act. Acc                       \\ 
Dou-no-prob               & 1,500                                         & 1,712,802                                     & 0.29 T                                  & CoT RLsum, Act. Acc.       \\ 
Dou-pred-prob            & 2000                        & 19,422                        & N/A                 & Pred. Acc        \\ 
Go-next-state          & N/A                                           & 150,000                                       & 0.23 T                                  & s' Acc.                        \\ 
Go-analysis          & 36                                            & 138,693                                       & 0.22 T                                  & Score/Winrate MAE         \\ 
Go-state-expl        & N/A                                           & 1,503                                         & 0.01 T                                  & expl. RLsum/ppl.        \\ \hline

\end{tabular}
\label{table:dataset_info}
\caption{Overview of datasets used for Doudizhu and Go tasks.}
\end{table}

\begin{figure}[!h]
\centering
\includegraphics[width=0.68\linewidth]{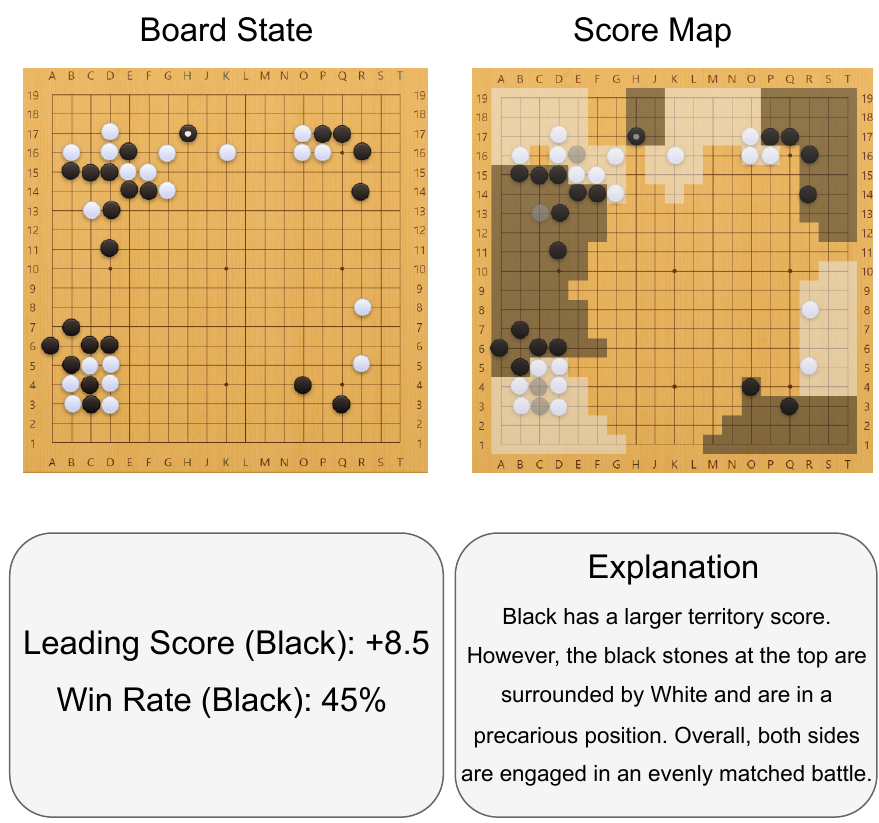}
\caption{An example for step-by-step reasoning in Go.}
\label{fig:go-rule}
\end{figure}

\section{Step-by-step Reasoning Process while Playing Go}
\label{appendix:go-rule}

In this section, we outline the step-by-step reasoning process in Go by analyzing key aspects of the game state. Effective decision-making in Go involves evaluating multiple factors, such as the current board position, ownership of intersections, territorial balance, and strategic threats. Figure~\ref{fig:go-rule} illustrates this process visually.

\begin{itemize}
    \item The top-left diagram shows the current state of the board, including all past moves made by both black and white players.
    \item The top-right diagram shows the ownership status of each intersection on the board (occupied by black, white, or currently undetermined). It is important to note that even if an intersection is currently occupied by one side, it could potentially be reclaimed by the opponent in subsequent moves. This task requires a deep understanding of specific board areas, which may span multiple lines, in order to be executed accurately.
    \item The bottom-left diagram presents the current difference in the number of territories held by black and white, as well as the win probability for black.
    \item The bottom-right diagram contains a commentary on the current game situation. As can be seen, although black currently leads in territory, the win probability slightly favors white. This is because the black stones in the upper area are surrounded by white, giving white a potential opportunity to gain an advantage. Therefore, evaluating the win probability is a complex process that involves taking multiple factors into account.
\end{itemize}

\section{The Correlation Examples between the Strategy in Doudizhu and the Opponent's Strategy}
\label{appendix:dou_example}
In fact, in Doudizhu, the optimal card-playing strategy is highly correlated with the opponent's strategy. Here is a simple example for illustration. Suppose the rule for the card size follows the basic rules of "Fight the Landlord", but to simplify the scenario, let's assume there are only two players, with the following cards in their hands:
My cards: Q, A, 2; Opponent's cards: K, 2, 2

Now, I play first. Suppose the opponent's strategy is: always play the smallest legal card, then one of my optimal strategies would be: Q (me) - K (opponent) - 2 (me) - PASS (opponent) - 2 (me) - I win.
However, if the opponent's strategy changes to: always play the largest legal card, it would go: Q (me) - 2 (opponent) - PASS (me) - 2 (opponent) - PASS (me) - K (opponent) - I lose.

Therefore, the optimal strategy for playing cards is related to the strategy adopted by the opponent.

\section{Training Hyper-parameters}
\label{appendix:hyper}

We list important training hyper-parameters in Table~\ref{tab:hyperparams}. The AdamW optimizer was used for improved weight decay handling, with a cosine learning rate scheduler to facilitate smooth convergence. Gradient clipping was applied to mitigate exploding gradients, and a warm-up phase of 3000 iterations was implemented to stabilize early training dynamics. Additionally, LoRA parameters were set to enhance model efficiency while maintaining performance, and a dropout rate of 0.1 was used to prevent overfitting.

\begin{table}[h]
    \centering
    \begin{tabular}{l|c|l|c}
        \hline
        \textbf{Hyper-Parameter} & \textbf{Value} & \textbf{Hyper-Parameter} & \textbf{Value} \\
        \hline
        Optimizer & AdamW & LR Scheduler & cos \\
        Learning Rate & 5e-5 & Gradient Clipping Norm & 1.0 \\
        Beta1 & 0.99 & Warm-up Iterations & 3000 \\
        Beta2 & 0.999 & LoRA-r & 32 \\
        Weight Decay & 0.01 & LoRA-$\alpha$ & 64 \\
        Batch Size & 8 & Dropout & 0.1 \\
        \hline
    \end{tabular}
    \caption{Hyper-parameters and their values}
    \label{tab:hyperparams}
\end{table}

\subsection{Discussion}
\label{appendix:discussion}
Concurrently, some recent groundbreaking models, such as LLaMA \citep{touvron2023llama} and Deepseek \citep{shao2024deepseekmath}, owe a significant part of their progresses to the integration of code and math problems during training. These works argue that leveraging code and math data improves the intrinsic reasoning skills of foundation models. This claim is also supported by subsequent experiments showing that CoT further enhances this advantage \citep{fu2022gptroadmap, wei2022chain}.

Several insights help explain this phenomenon: 1) Pre-training provides LLMs with exposure to the logic inherent in programming and mathematical languages, thus facilitating the development of systematic reasoning \citep{ma2023training}. 2) The dependencies and logic flow within these contents  contribute to the model's capabilities on capturing long-term dependency \citep{wang2022self, ma2023training}. 3) The deterministic nature of code execution and math derivation ensures non-duality in output for the same input, mitigating ambiguity in LLMs' logical deductions. 

We posit that data generated from decision-making games shares similar advantages, while introducing unique challenges such as \textit{uncertainty} and \textit{imperfect information}. Our experiments reveal that such data contributes to improving general reasoning abilities. Consequently, we propose that similar synthetic data generation strategies be incorporated into future pre-training phases of large models to enhance the logical structure and diversity of reasoning challenges in the training dataset.

\section{Case Study}
\label{appendix:case_study}

\subsection{Doudizhu Cases}
\begin{figure}[tb]
\centering
\includegraphics[width=1\linewidth]{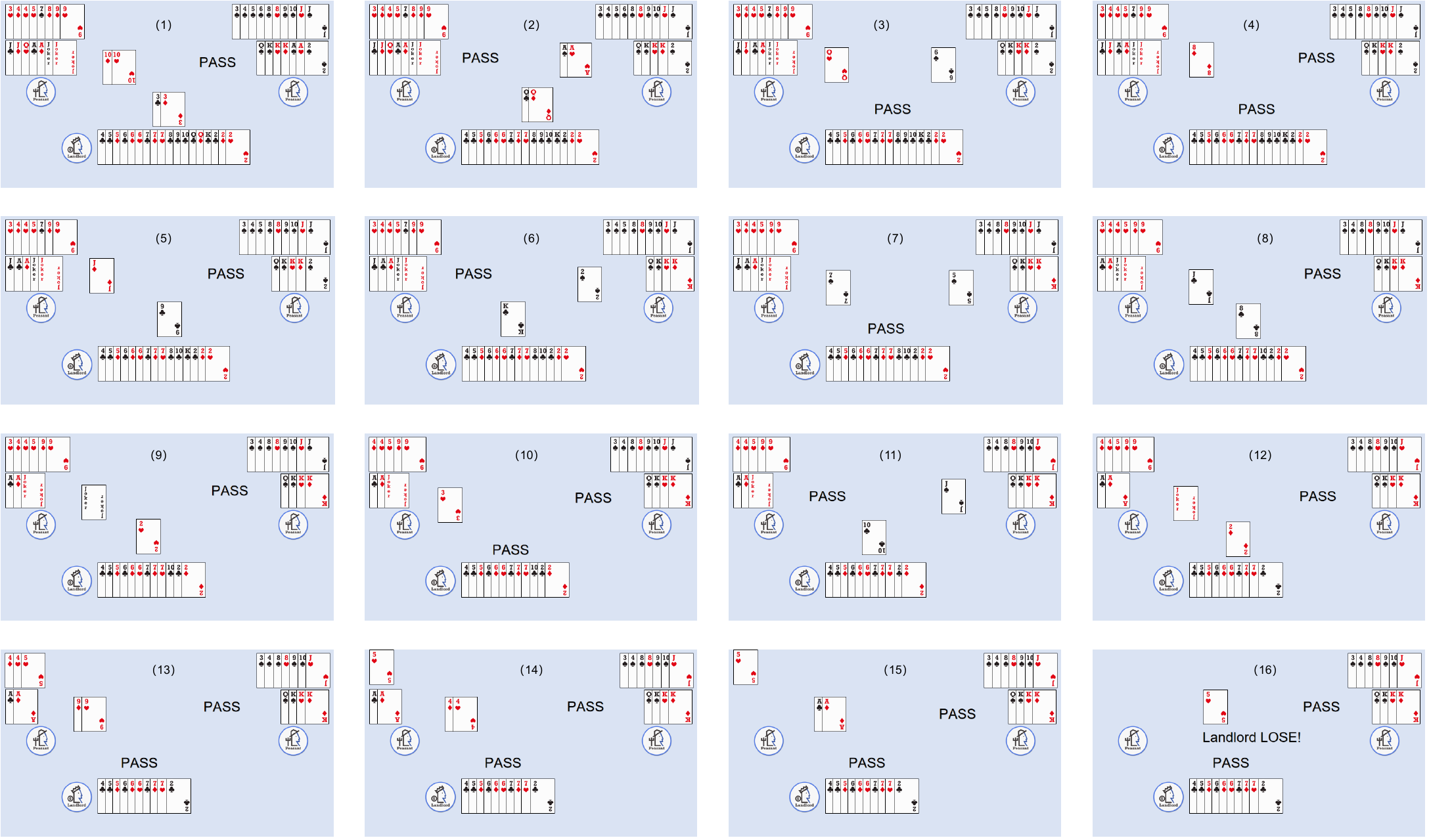}
\caption{Doudizhu game replay for DouZero (landlord) v.s. DouZero (peasants). Here is the key frames of a complete game. The peasants wins finally.}
\label{fig:dou_lose}
\end{figure}

\begin{figure}[tb]
\centering
\includegraphics[width=1\linewidth]{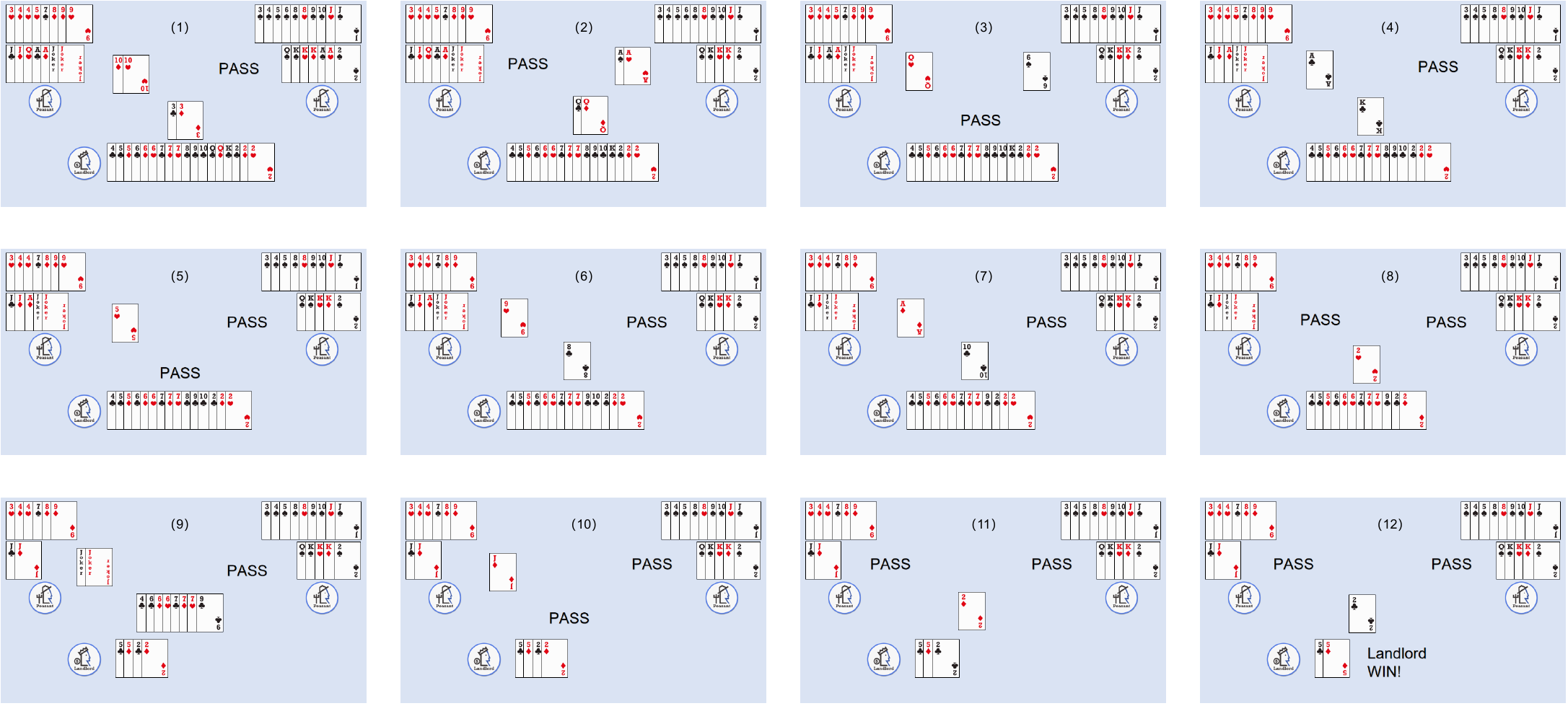}
\caption{Doudizhu game replay for Mastermind (landlord) v.s. DouZero (peasants). The landlord wins by playing more strategically.}
\label{fig:dou_win}
\end{figure}

In this section, we analyze several examples to demonstrate that MasterMind has effectively mastered the game of Doudizhu. Specifically, we aim to show that MasterMind does not merely memorize expert move sequences but instead learns the reasoning process, allowing it to generalize its knowledge to unseen game situations. In some cases, it can even match or surpass expert performance.  
To illustrate this, we compare the performance of MasterMind and the expert model DouZero as the landlord, using the same initial hand setup in Fig.~\ref{fig:dou_lose} and Fig.~\ref{fig:dou_win}. From the third round onward, MasterMind adopts a different strategy from the expert, playing more aggressively to prevent the peasants from making moves. As a result, MasterMind wins the game, whereas DouZero fails. This outcome strongly indicates that MasterMind does not rely on fixed move patterns but instead engages in strategic reasoning.

\subsection{Go Cases}
\begin{figure}[tb]
\centering
\includegraphics[width=0.9\linewidth]{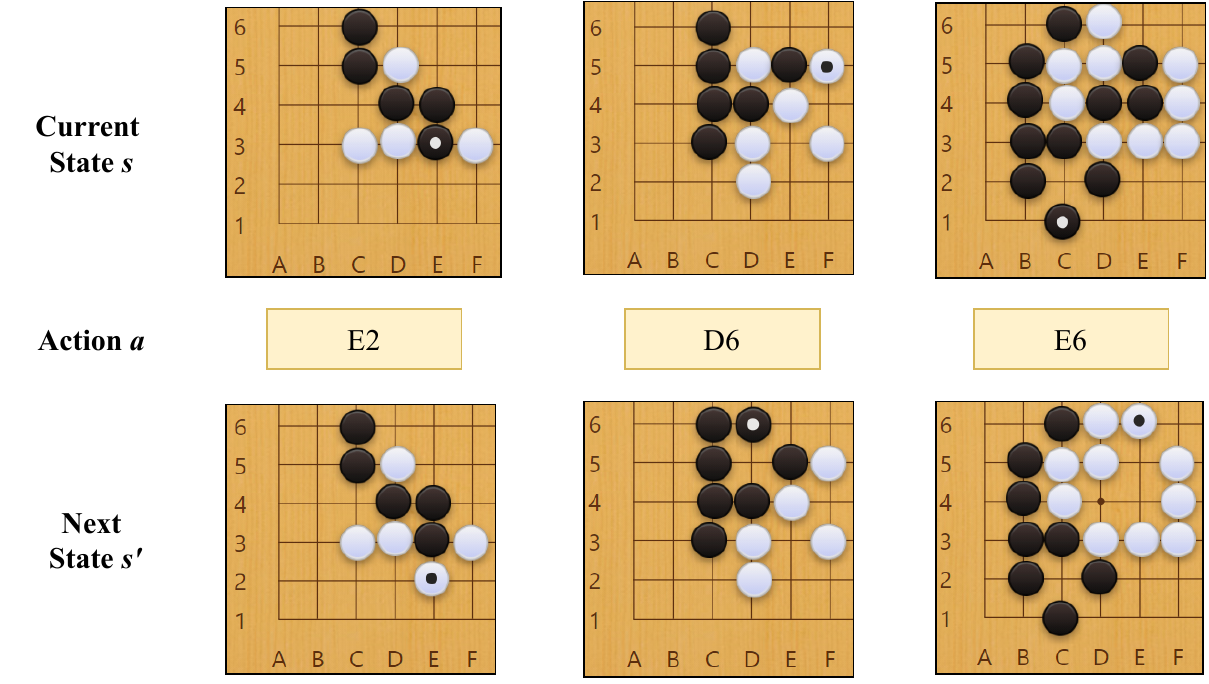}
\caption{Data examples about $s, a\rightarrow s'$ for Go.}
\label{fig:go_example}
\end{figure}

Fig.~\ref{fig:go_example} presents examples of predicting the next board state in Go based on the current board configuration and the move played. The difficulty of prediction increases progressively from left to right across the three examples.  
In the leftmost example, the board change is minimal, as it results solely from the placement of a new stone. In the middle example, an opponent's stone is captured and removed, meaning the move not only affects its immediate location but also alters other positions on the board. In the rightmost example, multiple opponent stones spanning several rows and columns are captured and removed, highlighting the complexity of Go’s rules.  

Notably, MasterMind accurately predicts board state transitions in all three scenarios and generalizes well to unseen states, demonstrating its strong understanding of Go’s mechanics.

\begin{figure}[tb]
\centering
\includegraphics[width=0.8\linewidth]{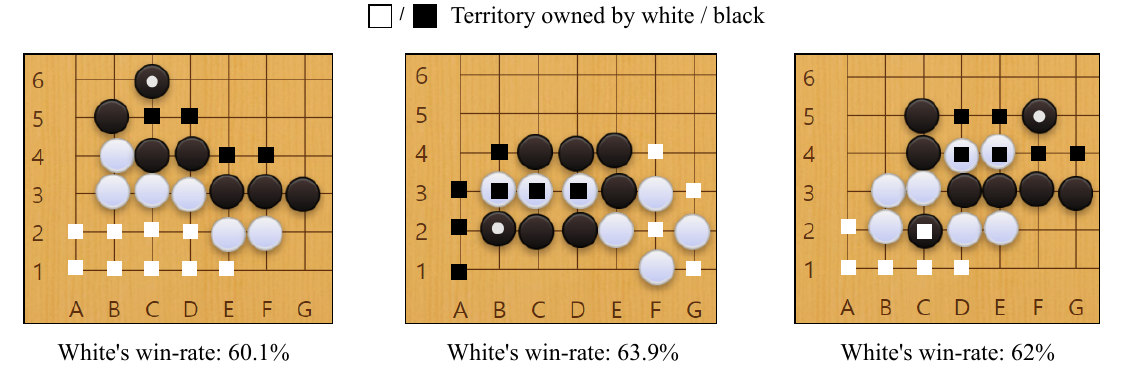}
\caption{Data examples about territiries and win-rate prediction.}
\label{fig:go_example2}
\end{figure}

Fig.~\ref{fig:go_example2} presents several examples of territory estimation and win rate prediction in Go.  In the leftmost example, MasterMind accurately identifies that White has secured territory in the corner by leveraging the edge of the board. Given that the model processes the board as a one-dimensional sequence, this demonstrates its ability to analyze territory across multiple rows, highlighting its strong long-sequence modeling capability.  

In the two rightmost examples, some of White's stones are completely surrounded by Black. While they have not yet been captured, they are effectively trapped with no escape. MasterMind correctly recognizes this situation, which requires an understanding of the board edges and spatial relationships spanning multiple rows—again showcasing its advanced long-sequence reasoning ability.  

Regarding win rate prediction, although White does not lose any stones in the leftmost example, its overall win rate is lower than in the other two cases, where some stones have already been captured. This aligns with the complexity of win rate estimation in Go, which depends not only on the current board state but also on future potential developments. MasterMind demonstrates a reasonable grasp of this intricate evaluation process.

\end{document}